\documentclass[10pt,twocolumn,letterpaper]{article}

\usepackage[pagenumbers]{cvpr} % To force page numbers, e.g. for an arXiv version

\definecolor{cvprblue}{rgb}{0.21,0.49,0.74}
\usepackage[pagebackref,breaklinks,colorlinks,allcolors=cvprblue]{hyperref}
\usepackage{algorithmic}
\usepackage{algorithm}
\usepackage{multirow}
\usepackage{array}
\usepackage{tabularx}
\usepackage{color}
\def\paperID{12673} % *** Enter the Paper ID here
\def\confName{CVPR}
\def\confYear{2026}

\title{HeatV2X: Scalable Heterogeneous Collaborative Perception via Efficient Alignment and Interaction}

\author{Yueran Zhao\\
{\tt\small zhaoyueran@bit.edu.cn}
\and
Zhang Zhang\\
{\tt\small zhangzhang00@bit.edu.cn}
\and
Chao Sun\\
{\tt\small chaosun@bit.edu.cn}
\and
Tianze Wang\\
{\tt\small 3120230483@bit.edu.cn}
\and
Chao Yue\\
{\tt\small 3120235513@bit.edu.cn}
\and
Nuoran Li\\
{\tt\small 3220230398@bit.edu.cn}
\thanks{All authors are with Beijing Institute of Technology, Beijing, China.}
}

\begin{document}
\maketitle
\begin{abstract}
% The ABSTRACT is to be in fully justified italicized text, at the top of the left-hand column, below the author and affiliation information.
% Use the word ``Abstract'' as the title, in 12-point Times, boldface type, centered relative to the column, initially capitalized.
% The abstract is to be in 10-point, single-spaced type.
% Leave two blank lines after the Abstract, then begin the main text.
% Look at previous \confName abstracts to get a feel for style and length.
Vehicle-to-Everything (V2X) collaborative perception extends sensing beyond single vehicle limits through transmission. However, as more agents participate, existing frameworks face two key challenges: (1) the participating agents are inherently multi-modal and heterogeneous, and (2) the collaborative framework must be scalable to accommodate new agents. The former requires effective cross-agent feature alignment to mitigate heterogeneity loss, while the latter renders full-parameter training impractical, highlighting the importance of scalable adaptation. To address these issues, we propose \textbf{He}terogeneous \textbf{A}dap\textbf{t}ation  (HeatV2X), a scalable collaborative framework. We first train a high-performance agent based on heterogeneous graph attention as the foundation for collaborative learning. Then, we design Local Heterogeneous Fine-Tuning and Global Collaborative Fine-Tuning to achieve effective alignment and interaction among heterogeneous agents. The former efficiently extracts modality-specific differences using Hetero-Aware Adapters, while the latter employs the Multi-Cognitive Adapter to enhance cross-agent collaboration and fully exploit the fusion potential. These designs enable substantial performance improvement of the collaborative framework with minimal training cost. We evaluate our approach on the OPV2V-H and DAIR-V2X datasets. Experimental results demonstrate that our method achieves superior perception performance with significantly reduced training overhead, outperforming existing state-of-the-art approaches. Our implementation will be released soon.
\end{abstract}
\section{Introduction}
\label{sec:intro}

\begin{figure*}[!htbp]
\centering
\includegraphics[width=\textwidth]{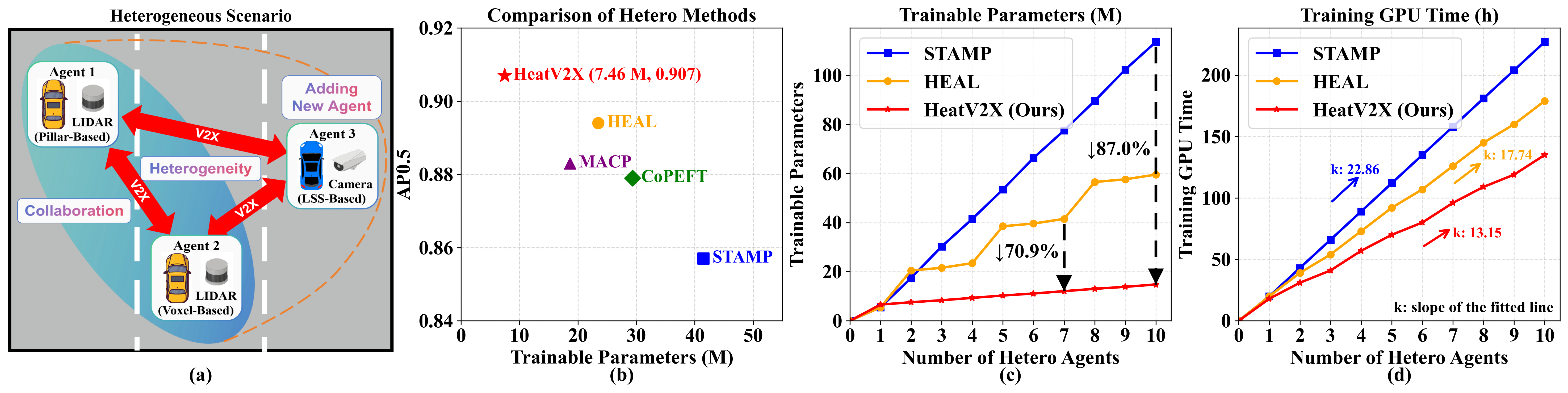}
\caption{\textbf{Comparison with heterogeneous collaborative perception methods on OPV2V-H.} (a) Typical heterogeneous collaboration scenarios and expansion processes. (b) Comparison of heterogeneous methods in terms of perception performance and model efficiency with 4 agents. (c) and (d) The trainable parameters and training GPU time versus the number of heterogeneous agents. \textbf{Please note that the pre-training of each modality is also included in the calculation.} }
\label{fig:hetero_and_scalable}
\end{figure*}

Multi-agent collaborative perception leverages information sharing across multiple viewpoints to achieve more accurate and comprehensive perception. Agents from vehicles and roadside infrastructure communicate critical information via V2X \cite{yu2024_univ2x,zhang2025height3d}, integrating individual perceptions into a unified whole that enables blind-spot compensation and beyond-line-of-sight perception \cite{10979246}. However, agents exhibit multi-modal heterogeneity in aspects such as input modalities, network architectures, and communication latency, which inevitably poses significant challenges to collaborative perception \cite{Gao_2025_CVPR}.

The problem of multi-modal heterogeneity arises from discrepancies among agents, such as modality heterogeneity and model heterogeneity \cite{li2024delving}. A critical challenge in heterogeneous collaborative perception lies in whether the fusion process can effectively account for and adapt to the characteristics of each agent. Late fusion, which shares and aggregates detection results, can mitigate modality heterogeneity to some extent. However, variations in perception accuracy across different network models often lead to a confidence crisis. Early fusion ensures that data are integrated within a single agent’s network, thereby avoiding model heterogeneity. Yet, due to bandwidth disparities among agents, full-scale data transmission is impractical. Recent advanced works have explored feature-level fusion through feature sharing \cite{Where2comm:22,10791908,qu2023sicp,Xiang_2023_ICCV,YueCodeFilling:CVPR2024,zhou2025pragmatic}, where end-to-end joint training enables adaptation to multi-modal data and heterogeneous models. While this approach partially alleviates heterogeneity, it essentially embeds potential compensation for heterogeneous agents into the network parameters as a black-box process. Such training, however, is fragile: once agents are replaced, the existing collaborative perception framework may no longer remain applicable.

\begin{figure*}[!htbp]
\centering
\includegraphics[width=\textwidth]{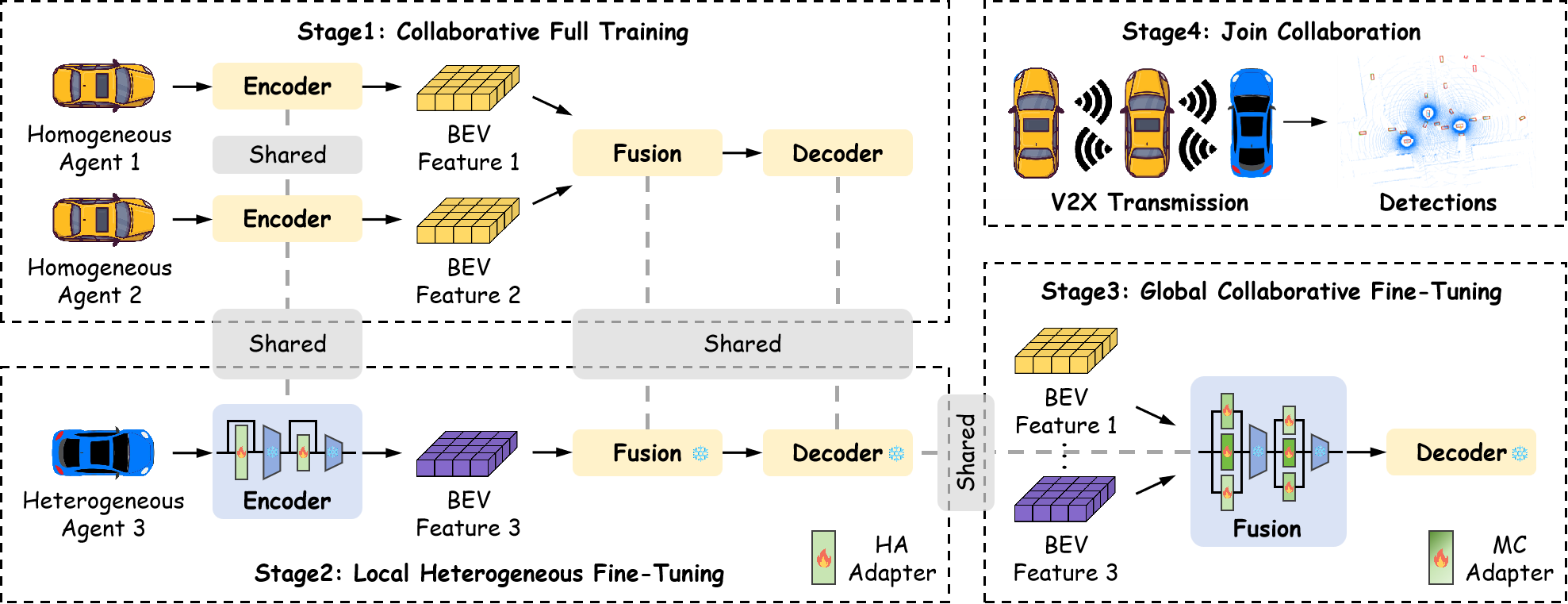}
\caption{\textbf{Overview of HeatV2X.} Stage 1: We obtain the collaborative base model via full-parameter training. Stage 2: The new agent freezes most of the weights shared from the base and performs rapid alignment using the Hetero-Aware Adapters (HA Adapter). Stage 3: The collaboration conducts cross-agent interaction based on the Multi-Cognitive Adapter (MC Adapter). Stage 4: Collaborative perception.}
\label{fig:structure}
\end{figure*}

As neural networks continue to evolve, existing collaborative frameworks are constantly challenged by the integration of new heterogeneous agents, making repeated end-to-end full retraining prohibitively expensive. Consequently, shifting from full-parameter training to parameter-efficient fine-tuning has become a key trend in collaborative perception. The core idea is to update only a subset of parameters or newly introduced modules while achieving performance comparable to full fine-tuning\cite{han2024peft}. Representative studies such as \cite{lu2024an,gao2025stamp,zhou2025pragmatic} focus on partially training heterogeneous modules to align features to the primary agent’s space for collaborative fusion. Meanwhile, works like \cite{ma2024macp,wei2025copeft} adopt parameter-efficient fine-tuning to achieve cross-agent feature alignment. Compared to full training, these methods not only preserve or enhance perception performance, but also improve scalability and reduce model complexity. 

% However, most existing frameworks still perform one-way knowledge transfer—from the main agent to new agents—and seldom explore whether newly joined agents can, in turn, reinforce the collaboration itself.
However, current approaches such as HEAL \cite{lu2024an} mostly adopt a unidirectional interaction pattern, where the main collaborative framework transmits knowledge to newly joined agents. These new agents receive shared weights from the base model and perform alignment autonomously. In this process, the potential feedback or compensation effect of new agents toward the collaborative fusion module is neglected. Although this design helps reduce the training cost, it also fails to explore whether newly joined agents can, in turn, reinforce the collaboration itself. In addition, methods with adaptation like CoPEFT \cite{wei2025copeft} and MACP \cite{ma2024macp} primarily focus on single modality without considering modality or encoder heterogeneity. Furthermore, the fine-tuning adapters used in STAMP \cite{gao2025stamp} are fully identical across agents, which may leads to feature homogenization and heterogeneous loss for agents without pretraining.

To address the challenges above, we propose \textbf{Heterogeneous Adaptation  (HeatV2X)},  a scalable collaborative perception framework that enables multilateral collaboration and rapid adaptation among heterogeneous models through local and global fine-tuning. We perform full-parameter training to obtain a high-performance agent as the ego agent, and broadcast its pose features and model parameters to other agents. Subsequently, we perform Local Heterogeneous Fine-Tuning (LHFT) and Global Collaborative Fine-Tuning (GCFT) with two specialized types of adapters. During the LHFT stage, we employ Hetero-Aware Adapters (HA Adapter) with specific convolution and insert it progressively across layers to extract multi-level heterogeneous features while maintaining a low parameter budget. But the vanilla adapters fail to model complex collaboration when it comes to the GCFT stage. Therefore, we introduce the Multi-Cognitive Adapter (MC Adapter), which captures cross-agent interactions through multi-scale convolutions. The MC Adapter is integrated into the fusion module via a residual connection to support global collaborative fine-tuning. During training, each new agent first performs local heterogeneous fine-tuning to map its features into a unified representation space, and then broadcasts the aligned features and local parameters back to the main agent. The framework subsequently performs global collaborative fine-tuning to strengthen cross-agent interactions and further exploit the potential of collaborative perception. Throughout this process, heterogeneous information and collaborative information are stored separately in different adapters, enabling low-parameter heterogeneous alignment and collaboration enhancement.

To validate the performance of HeatV2X against baseline methods, we conduct experiments on the OPV2V-H and DAIR-V2X datasets. Our comparison goes beyond model performance to include trainable parameters and training GPU time, with additional noise and latency introduced to evaluate the model’s robustness.
As shown in~\cref{fig:hetero_and_scalable}, our method shows superiority in performance, trainable parameter efficiency, and scalability. Each time a completely untrained agent is introduced, our method achieves adaptation with minimal trainable parameters and GPU hours. The results demonstrate that HeatV2X effectively addresses the challenges of multi-modal collaboration, achieving steady improvements in joint perception performance as heterogeneous agents are incrementally integrated. Moreover, HeatV2X surpasses state-of-the-art (SOTA) collaborative perception methods on 3D object detection. 
Our main contributions are summarized as follows:

\begin{itemize}

\item{We propose a scalable collaborative perception framework, Heterogeneous Adaptation (HeatV2X), which enables rapid alignment for newly participating heterogeneous agents and further unlocks their collaborative potential.}

\item{To alleviate the heavy training burden when incorporating new agents, we propose a Local Heterogeneous Fine-Tuning (LHFT) method. It employs Hetero-Aware Adapters (HA Adapter) for heterogeneous encoders to efficiently capture and retain modality-specific information without imposing a high parameter burden.}

\item{To investigate the potential gains of newly added agents for the collaborative fusion module, we introduce a Global Collaborative Fine-Tuning (GCFT) strategy. During this process, the Multi-Cognitive Adapter (MC Adapter) enables cross-agent interaction, efficiently enhancing collaborative fusion performance with minimal parameters overhead.}

\item{Extensive experiments demonstrate that HeatV2X outperforms existing SOTA methods in object detection, significantly improving robustness and scalability in collaborative perception at a low computational cost.}

\end{itemize}

\section{Background}
\label{sec:Background}

\subsection{V2X Heterogeneous Perception}

\subsubsection{Multi-Agent Heterogeneous}
Early collaborative frameworks primarily explored single-modality fusion across multiple agents, where inter-agent variation arose mainly from differences in viewpoint or pose \cite{zhang2025heightformer,zhang2025pillarmamba}, while the input modality and feature encoders remained largely identical. For instance, V2VNet \cite{V2VNet} employs a spatially aware graph neural network (GNN) to aggregate LiDAR point-cloud features from multiple agents. Similarly operating on point clouds, DiscoNet \cite{NEURIPS2021_f702defb} introduces a knowledge-distillation scheme that aligns the student model’s post-collaboration feature maps with those of a teacher model to facilitate the construction of a collaborative student network. For multi-view imagery, CoBEVT \cite{xu2022cobevt} proposes a fused axial-attention mechanism to integrate camera features from different viewpoints and agents. To achieve communication-efficient and robust perception, ERMVP \cite{Zhang_2024_CVPR} further adopts a filtered-and-merged feature sampling strategy and adds a feature-space calibration module to align agent poses.

\subsubsection{Multi-Modal Heterogeneous}
As the number of collaborating agents increases, V2X inevitably faces the challenge of cross-modal fusion. The pioneering V2X-ViT \cite{xu2022v2xvit} captures inter-agent spatial interactions by performing heterogeneous multi-agent self-attention with multi-scale windows.
Building on this, HM-ViT \cite{Xiang_2023_ICCV} introduces both global and local heterogeneous 3D attention modules to further exploit cross-agent information. A similar global–local design is adopted by V2VFormer++ \cite{10265751}, but with a distinct dynamic channel fusion mechanism for adaptive cross-modal aggregation. SparseAlign \cite{Yuan_2025_CVPR} addresses the center-feature missing (CFM) and isolated convolution field (ICF) problems in sparse convolution by employing coordinate-enhanced convolution (CEC). 

Although these methods effectively explore collaborative domain adaptation, they incur high training costs and require retraining whenever the set of collaborating agents changes, revealing limited robustness.

\subsection{Fine-Tuning Strategy}
\subsubsection{Parameter-Efficient Fine-Tuning}
Multi-modal heterogeneous collaborative frameworks typically require additional parameters for each newly added agent, making full-scale training impractical for resource-constrained participants. As a result, cost-efficient strategies for collaborative domain adaptation have become a central focus in heterogeneous collaborative perception.

Parameter-efficient fine-tuning methods have achieved remarkable success in image vision tasks and large-scale models. Techniques such as Adapter \cite{adapter}, BitFit \cite{bitfit}, LoRA \cite{hu2022lora}, and Mona \cite{yin20255} update only a small set of newly introduced or layer-specific parameters, significantly reducing computational cost while preserving the performance of pretrained models.

\subsubsection{Collaborative Perception Based on Fine-Tuning}
Recent studies have explored parameter-efficient strategies for heterogeneous encoders. MPDA \cite{xu2023mpda} employs a learnable feature adaptor together with a Sparse Cross-domain Transformer to achieve feature alignment across heterogeneous agents. MACP \cite{ma2024macp} trains only the convolutional adapters (ConAda) within heterogeneous encoders and applies scaled and shifted features (SSF) for cross-agent feature alignment. CoPEFT \cite{wei2025copeft} performs parameter-efficient fine-tuning directly on the shared feature maps of all agents, enabling rapid adaptation of pretrained collaborative perception models to new deployment environments.

HEAL \cite{lu2024an} is a pioneering work in this direction. In this work, a high-performance model parameters of the collaborative backbone are shared with newly joined agents. It aligns heterogeneous agents to a unified feature space by training only the encoder and employing a backward-alignment mechanism. Building on this idea, PnPDA \cite{pnpda} adopts a shared standard semantic space as an intermediate representation: features from heterogeneous vehicles are first mapped to this shared space and then projected by the ego vehicle into its own semantic space. Similarly, STAMP \cite{gao2025stamp} further freezes the encoder and aligns heterogeneous features by retraining only agent-specific adapter–reverter pairs to map them into a unified semantic space.

However, neither the further reduction of trainable parameters nor the enhancement of fusion via agent interactions has been investigated. Motivated by this, we propose HeatV2X, which performs bidirectional interaction among collaborating vehicles to enhance the generalization and perception performance of the collaborative framework. Meanwhile, lightweight Adapter-based fine-tuning replaces full-parameter and encoder training, further reducing model complexity and training cost.
\section{Methodology}
\label{sec:Methodology}

\subsection{Collaborative Base Model}

Our collaborative framework performs feature-level fusion. Agents aligned within the unified collaboration system are regarded as the ego modality, and all features are projected into the collaborative feature space. For the $i$-th agent, the raw observation (point cloud or image) $O_i$ is processed by the encoder $f_{encoder}$ to extract its ego-space feature map $F_i$. The feature map is then passed through the transmission module $f_{trans}$, where both $F_i$ and the relative pose information $M_{ij}$ are shared with agent $j$ for subsequent fusion. The fusion module performs feature-level aggregation to produce the fused representation $H_i$, which is further processed by the detection head $f_{head}$ to generate collaborative detection results.

% \subsubsection{The Heterogeneous Encoder}
For each raw input $O_i$, a heterogeneous encoder ($\text{Encoder}_H$) extracts modality-specific features, which are then processed by a unified backbone network composed of several convolutional blocks for multi-scale feature extraction. A subsequent dimensional alignment maps all features into a unified 256-dimensional space to ensure structural consistency across modalities. The process is formulated as follows:

\begin{equation}
\label{eq:encoder}
  F_{i} =\text{Shrink}(\text{Bacbone}( \text{Encoder}_{H}(O_i)))
\end{equation}
where $\text{Encoder}_{H}$ denotes the heterogeneous encoder, $F_i$ the corresponding feature map, and $\text{Shrink}$ the dimensionality reduction module for efficient fusion.

% \subsubsection{Heterogeneous Vison Transformer}
% \subsubsection{Vison Transformer}

Our fusion module is built upon a vision transformer. The input feature representation is denoted as $F_i \in \mathbb{R}^{B\times L\times H\times W\times(C+3)}$, where $B$ is the batch size, $L$ the number of agents, $H$ and $W$ the spatial dimensions, $C$ the feature dimension, and the additional three channels encode spatial positional information.

We first perform spatio-temporal alignment to transform features from other agents into the ego feature space, obtaining $F_{i\rightarrow j}$. The aligned features from all agents are then concatenated along the channel dimension and linearly projected to form $F_{j}^{(N)} \in \mathbb{R}^{N\times H\times W\times C}$, which serves as the input to the subsequent attention module.

% \begin{equation}
% \label{eq:project}
%   F_{j}^{(N)} = \text{Concat}(f_{\text{project}_{n\rightarrow j}}(F_n)), n=0,1,...,N
% \end{equation}
% where $n$ refer to the number of the agents and $F_{j}^{(N)}$ is subsequently used as the input for the attention module.

% Subsequently, $F_{i\rightarrow j}^{(1)}$ is linearly projected into $Q_{ij}$, $K_{ij}$, and $V_{ij}$ representations:

% \begin{equation}
% \label{eq:qkv_linear}
% \{Q_{ij}, K_{ij}, V_{ij}\} = \{W_q, W_k, W_v\} \cdot F_{i\rightarrow j}^{(1)}.
% \end{equation}
% where $W_q$, $W_k$, and $W_v$ are learnable project weight matrices. Cross-attention is then computed as

Subsequently, the aligned feature $F_{i\rightarrow j}^{(1)}$ is linearly projected into $Q_{ij}$, $K_{ij}$, and $V_{ij}$ representations using learnable matrices $W_q$, $W_k$, and $W_v$. Cross-attention is then applied to compute feature interactions between the ego agent $j$ and other agents $i$ and $k$.

\begin{equation}
\label{eq:att}
  F_{ik}^j =  \text{softmax}\left(\frac{Q_{ij}K_{kj}^T}{\sqrt{d_k}}\right)V_{kj}
\end{equation}
where $j$ denotes the ego agent, and $i,k$ indicate the sources of $Q_{ij}$ and $K_{kj}/V_{kj}$, respectively.

% This unified formulation performs self-attention when $i=k$ and cross-attention when $i\neq k$, enabling the ego agent to incorporate contextual information from other agents.

\subsection{Collaborative Aligning and Fine-Tuning}

In collaborative frameworks such as HEAL\cite{lu2024an} and STAMP\cite{gao2025stamp}, the training of new agents relies on pre-training and reverse alignment to map their features into a unified collaborative space. These methods typically obtain a performance-degraded new agent through pre-training, followed by heterogeneous fusion with a reference model. This motivates us to explore whether parameter-efficient fine-tuning can be employed to independently align heterogeneous agents and the collaborative framework. Accordingly, we adopt the HA Adapter for fine-tuning new agents, and utilize the MC Adapter to optimize the fusion component.

\begin{figure}[!htbp]
\centering
\includegraphics[width=3.3in]{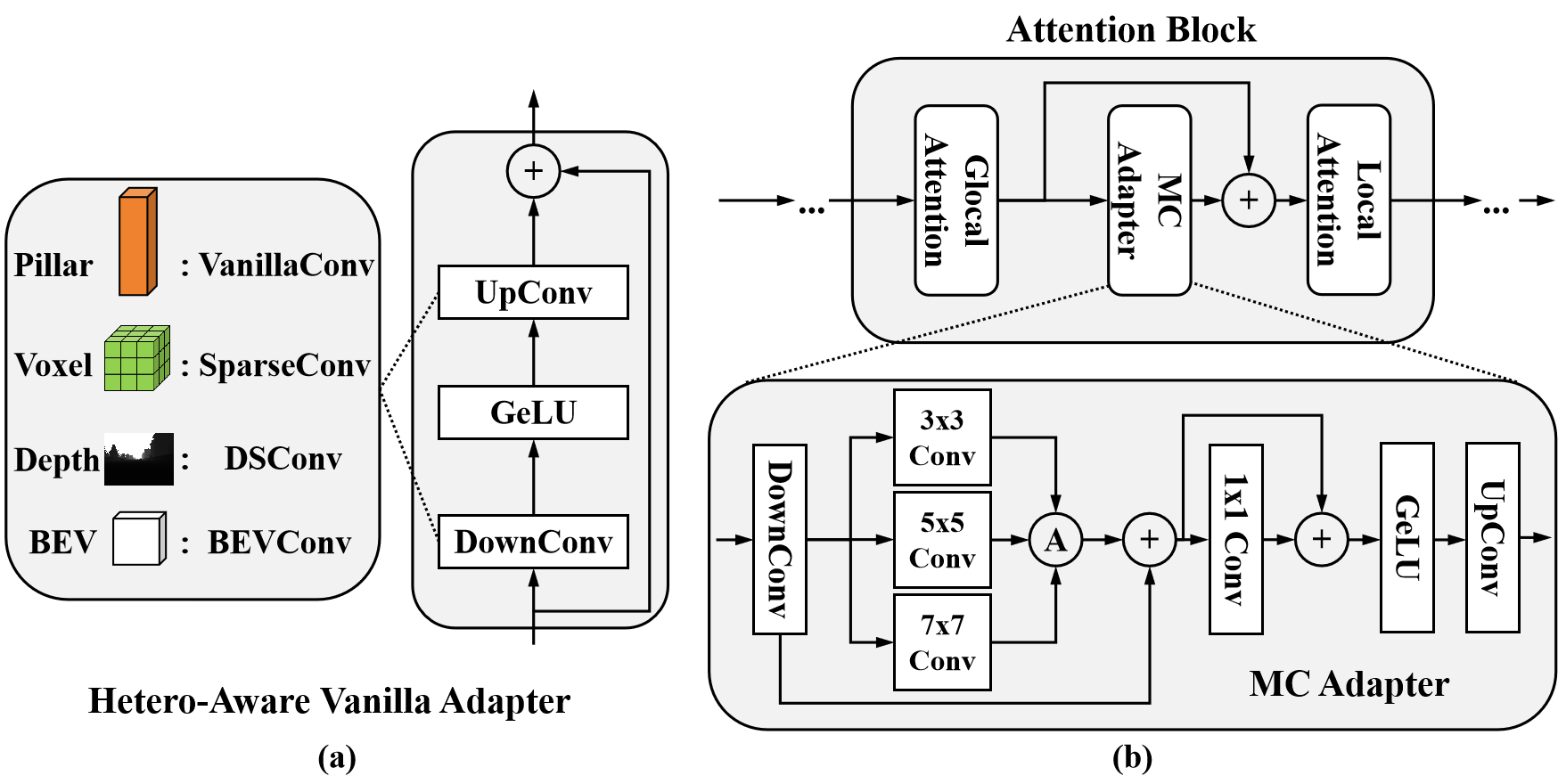}
\caption{\textbf{The structure of adapters.} (a): The Hetero-Aware Vanilla Adapter. (b): The MC Adapter.}
\label{fig:adapters}
\end{figure}

\subsubsection{Hetero-Aware Adapter (HA Adapter)}
The basic structure of our HA adapter is shown in~\cref{fig:adapters}. The adapter adopts different designs depending on the input modality. For encoders processing point cloud, we employ vanilla convolutions for PointPillars \cite{lang2019pointpillars} and sparse convolutions for SECOND \cite{yan2018second}. For depth image encoders, we adopt an adapter with depthwise separable convolutions (DSConv), thereby preserving the modality-specific characteristics of each encoder to the greatest extent. To further process the Bird’s Eye View (BEV) features, we integrate a BEV convolution module after the encoder. The above process can be formulated as follows: 

\begin{equation}
\label{eq:adapter}
  F_{ij}^i =  f_\text{Conv}^{\text{Up}}(f_\text{Activation}(f_\text{Conv}^{\text{Down}}(F_i)))
\end{equation}

\begin{equation}
\label{eq:heterconv}
	f_\text{Conv} = \begin{cases}
    \text{Conv}, &\text{if } x \in \mathbb{M}_\text{Pillar}\\
    \text{SparseConv}, &\text{if } x \in \mathbb{M}_\text{Voxel}\\
	\text{DSConv}, &\text{if } x \in \mathbb{M}_\text{Depth}\\
    \text{BEVConv}, &\text{if } x \in \mathbb{M}_\text{BEV}\\
		   \end{cases}
\end{equation}
where $\mathbb{M}x$ denotes the modality-specific input space, and $f\text{Conv}^{\text{Down}}$ and $f\text{Conv}^{\text{Up}}$ represent the downsampling and upsampling convolutions within the adapter.

\subsubsection{Local Heterogeneous Fine-Tuning (LHFT)}

As shown in \cref{fig:structure}, the Hetero-Aware Adapter is embedded between two downsampling layers and after the encoder to compensate for heterogeneous information lost during feature alignment. Unlike the vanilla adapter that processes all modalities identically, our design is modality-aware, enabling faster convergence and improved performance.

During the heterogeneous agents training stage, we freeze all components, including the MC Adapter that performs feature-level fusion, and only keep the heterogeneous adapters trainable. We adopt Focal Loss for classification $\mathcal{L}_{\text{c}}^{x}$, Smooth L1 Loss for regression $\mathcal{L}_{\text{r}}^{x}$, a direction loss $\mathcal{L}_{\text{d}}^{x}$ for orientation, and a depth loss $\mathcal{L}_{\text{depth}}^{x}$ for depth supervision. The unified objective for both single-agent and collaborative modes is

\begin{equation}
\mathcal{L} =
\lambda_{\text{c}} \mathcal{L}_{\text{c}}^{\text{single}}
+ \lambda_{\text{r}} \mathcal{L}_{\text{r}}^{\text{single}} \\
+ \lambda_{\text{d}} \mathcal{L}_{\text{d}}^{\text{single}} \\
+ \lambda_{\text{depth}} \mathcal{L}_{\text{depth}}^{\text{single}}
\end{equation}
where, the $\lambda_{\text{c}},\lambda_{\text{r}},\lambda_{\text{d}}$ and $\lambda_{\text{depth}}$ is the scaling factors. In the LHFT stage, the loss is calculated using the single-agent ground truth.

\subsubsection{Multi-Cognitive Adapter (MC Adapter)}
To further enhance feature extraction and facilitate parameter-efficient fine-tuning, we incorporate the Multi-Cognitive Adapter (MC Adapter) into the Fusion module.

% , as illustrated in the \cref{fig:havit}.
The vanilla adapter, as illustrated in~\cref{eq:adapter}, is relatively shallow, limiting its capacity for deeper feature interactions across heterogeneous modalities. In contrast, the MC Adapter incorporates multi-scale convolutions that capture features at diverse receptive fields, thereby providing richer and complementary information for the output.

Specifically, the upstream feature $F_{i\rightarrow j}^{(N)}$ is first processed by LayerNorm and Down-sampling, after which it is passed through three convolutional filters with kernel sizes $3\times3$, $5\times5$, and $7\times7$. The outputs of these convolutions are averaged, and a residual-like connection is applied to mitigate performance degradation and gradient vanishing. The resulting feature is then aggregated via a $1\times1$ convolution while preserving a residual connection. Finally, the feature is fed into an activation layer (we adopt the GeLU module here) and up-sampled to its original spatial resolution. The corresponding formulation is given as:

\vspace{-0.5cm}
\begin{gather}
\label{eq:hmcadapter}
  F_{a} = \text{DownConv}(\text{LayerNorm}(F_{i\rightarrow j}^{(N)}))  \\
  F_{b} =  \text{Conv}_\text{3x3}(F_{a}) + \text{Conv}_\text{5x5}(F_{a}) + \text{Conv}_\text{7x7}(F_{a})   \\
  F_{c} = \text{Conv}_\text{1x1}(\text{Avg}(F_{b})) \\
  \widetilde{F}_{i\rightarrow j}^{(N)} =  \text{UpConv}(\text{GeLU}(F_{c}))
\end{gather}

% \subsubsection{Multi-Cognitive Adapter for Framework}
\subsubsection{Global Collaborative Fine-Tuning (GCFT)}
Inspired by the advanced work Mona~\cite{yin20255}, we leverage the MC Adapter to fine-tune the fusion module. As shown in~\cref{fig:adapters}, the MC Adapter is positioned between the local and global attention modules. It preserves the original attention capability through a residual structure and further mines the perception potential overlooked by conventional attention modules.

After the new agent is well pretrained, the features  $F_i$  extracted by different agents are transmitted to the ego agent responsible for heterogeneous fusion via V2X communication and concatenated to form $F_{j}^{(N)}$. 
During the integration of the aligned agent into the collaborative framework, we freeze all parameters except those in the MC Adapter to enable efficient online fine-tuning. The MC Adapter employs multi-cognitive convolutions to capture collaborative information across multiple scales, thereby enabling efficient cross-agent fusion. Similarly, We adopt $\mathcal{L}_{\text{cls}}^{\text{collab}}$, $\mathcal{L}_{\text{reg}}^{\text{collab}}$, $\mathcal{L}_{\text{dir}}^{\text{collab}}$, and $\mathcal{L}_{\text{depth}}^{\text{collab}}$ as the loss functions for backpropagation and model optimization.

\newcolumntype{Y}{>{\centering\arraybackslash}X}

\begin{table*}[!htbp]
\centering
\caption{Performance comparison of different models on OPV2V-H. AP0.5 and AP0.7 are reported with different component combinations. Model parameters (in millions) are also shown. $A_x$ denotes different agents, each equipped with a distinct encoder ($m_x$).}
\label{tab:mainresults}
\setlength{\tabcolsep}{1.5pt}
\begin{tabular*}{\textwidth}{@{\extracolsep{\fill}}>{\centering\arraybackslash}m{3.5cm}*{9}{>{\centering\arraybackslash}m{1.2cm}}@{}}
\toprule
\multirow{2}{*}{\textbf{Model}} & \multicolumn{3}{c}{\textbf{AP0.5}} & \multicolumn{3}{c}{\textbf{AP0.7}} & \multicolumn{3}{c}{\textbf{Trainable Params (M)}} \\
\cmidrule(lr){2-4} \cmidrule(lr){5-7} \cmidrule(lr){8-10}
 % & +$C_E$ & +$L_S$ & +$C_R$ & +$C_E$ & +$L_S$ & +$C_R$ & +$C_E$ & +$L_S$ & +$C_R$ \\
  & +$A_2$ & +$A_3$ & +$A_4$ & +$A_2$ & +$A_3$ & +$A_4$ & +$A_2$ & +$A_3$ & +$A_4$ \\
\midrule
No Fusion (only $A_1$) & 0.788 & 0.788 & 0.788 & 0.658 & 0.658 & 0.658 & - & - & - \\
Late Fusion & 0.775 & 0.833 & 0.834 & 0.599 & 0.685 & 0.685 & 20.25 & 6.34 & 7.15 \\
\midrule
HM-ViT \cite{Xiang_2023_ICCV} & 0.813 & 0.871 & 0.876 & 0.646 & 0.743 & 0.755 & 47.71 & 65.08 & 83.34 \\
HEAL \cite{lu2024an} & 0.826 & 0.896 & 0.897 & 0.726 & \textbf{0.812} & \textbf{0.813} & 15.01 & 1.10 & 1.91 \\
MACP \cite{ma2024macp} & 0.809 & 0.887 & 0.883 & 0.639 & 0.754 & 0.746 & 15.60 & 16.71 & 18.65 \\
CoPEFT \cite{wei2025copeft} & 0.825 & 0.878 & 0.879 & 0.635 & 0.663 & 0.664 & 26.55 & 27.52 & 29.32 \\
STAMP \cite{gao2025stamp} & 0.796 & 0.863 & 0.857 & 0.720 & 0.801 & 0.803 & 11.98 & 12.81 & 11.24 \\
\midrule
HeatV2X (Ours) & \textbf{0.853} & \textbf{0.913} & \textbf{0.914} & \textbf{0.730} & 0.804 & 0.806 & \textbf{0.96} & \textbf{0.81} & \textbf{0.96} \\
\bottomrule
\end{tabular*}
\end{table*}

\begin{table}[htbp]
\centering
\caption{Performance comparison of different models under various modality combinations on DAIR-V2X.}
\label{tab:dairv2x_results}
\setlength{\tabcolsep}{2.5pt}
\begin{tabular*}{\columnwidth}{@{\extracolsep{\fill}}lcccccc@{}}
\toprule
% \multirow{2}{*}{Model} & \multicolumn{2}{c}{A1+A2} & \multicolumn{2}{c}{A1+A3} & \multicolumn{2}{c}{A1+A4} \\
\multirow{2}{*}{Model} & \multicolumn{2}{c}{$A_1+A_2$} & \multicolumn{2}{c}{$A_1+A_3$} & \multicolumn{2}{c}{$A_1+A_4$} \\
\cmidrule(lr){2-3} \cmidrule(lr){4-5} \cmidrule(lr){6-7}
 & AP0.5 & AP0.7 & AP0.5 & AP0.7 & AP0.5 & AP0.7 \\
\midrule
HM-ViT \cite{Xiang_2023_ICCV} & 0.619 & 0.423 & 0.630 & 0.434 & 0.623 & 0.455 \\
HEAL \cite{lu2024an}  & 0.618 & 0.456 & 0.537 & 0.464 & 0.547 & 0.464 \\
MACP \cite{ma2024macp}   & 0.579 & 0.350 & 0.637 & 0.431 & 0.585 & 0.421 \\
CoPEFT \cite{wei2025copeft} & 0.659 & 0.447 & 0.618 & 0.425 & 0.646 & 0.450 \\
\midrule
HeatV2X & \textbf{0.662} & \textbf{0.462} & \textbf{0.652} & \textbf{0.465} & \textbf{0.651} & \textbf{0.473} \\
\bottomrule
\end{tabular*}
\end{table}

\section{Experiment}
\label{sec:Experiment}

\subsection{Experiment Setup}
\subsubsection{Datasets}
We conduct experiments on two datasets: OPV2V-H\cite{lu2024an} and DAIR-V2X\cite{dair-v2x}.  
OPV2V-H is an extended version of the simulation dataset OPV2V \cite{Xu2021OPV2VAO}, augmented with data from a 32-channel LiDAR and four depth cameras. The original OPV2V \cite{Xu2021OPV2VAO}, collected in the CARLA simulator \cite{Dosovitskiy17}, contains 11k LiDAR frames, 44k RGB images, and 232 annotated bounding boxes for vehicle-to-vehicle (V2V) cooperative perception. OPV2V-H adds 37k depth images and 32-line LiDAR frames, enriching multi-modal information and providing a stronger basis for heterogeneous cooperative perception research.  

DAIR-V2X \cite{dair-v2x} is a real-world dataset comprising 39k RGB images and LiDAR point clouds, along with 464k annotated bounding boxes. Data are collected from both roadside units and vehicles equipped with LiDAR and RGB sensors, supporting heterogeneous cooperative perception across diverse spatial configurations.

\subsubsection{Implementation}
We configure four sensing modalities: Agent 1 ($A_1$) uses $m_1$ (PointPillars \cite{lang2019pointpillars}) as the ego agent, while Agent 2–4 ($A_2 -A_4$) use $m_2$ (LSS \cite{philion2020lift}, EfficientNet-b0 \cite{pmlr-v97-tan19a}), $m_3$ (SECOND \cite{yan2018second}), and $m_4$ (LSS \cite{philion2020lift}, ResNet50 \cite{resnet}), respectively. $m_1$ is first pretrained as the base collaborative model, and its parameters initialize the training of new agents. The agents are then sequentially integrated, yielding joint perception results with one to four participants.
%  ($A_1, A_1+A_2, A_1+A_2+A_3,A_1+A_2+A_3+A_4$)

Datasets partitioning aligns with agent modalities. Single-agent training uses only modality-specific data with ground-truth (GT) boxes visible to that agent, whereas multi-agent collaboration leverages all modal data with GT boxes defined by combined coverage. To ensure fair evaluation, the perception range is kept consistent with the baseline. For the OPV2V-H dataset, taking the ego agent ($A_1$) as reference, the perception range is set to $[-102.4m, 102.4m]$ along both the x and y axes, and $[-1m, 3m]$ along the z axis. For the DAIR-V2X dataset, the range is defined as $[-102.4m, 102.4m]$ in the x direction, $[-51.2m, 51.2m]$ in the y direction, and $[-3.5m, 1.5m]$ in the z direction.

\subsection{Performance Evaluation}
\subsubsection{Main Perception Performance}

\begin{figure}[!htbp]
  \centering
  % The first line
  \begin{minipage}[t]{0.48\linewidth}
    \centering
    \includegraphics[width=\linewidth]{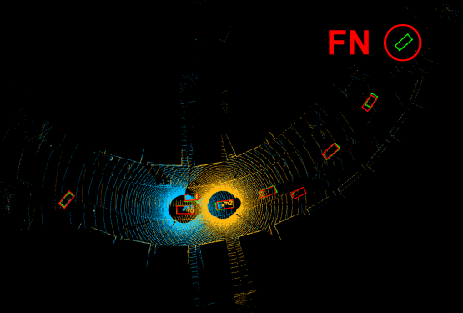}
    \subcaption{HEAL - Scene 1}
    \label{fig:vis_heal_1}
  \end{minipage}
  \hfill
  \begin{minipage}[t]{0.48\linewidth}
    \centering
    \includegraphics[width=\linewidth]{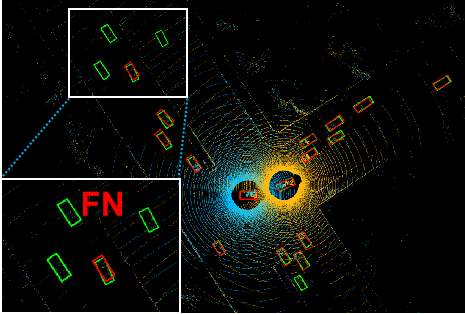}
    \subcaption{HEAL - Scene 2}
    \label{fig:vis_heal_2}
  \end{minipage}

  \vspace{0.8em}

  % The second line
  \begin{minipage}[t]{0.48\linewidth}
    \centering
    \includegraphics[width=\linewidth]{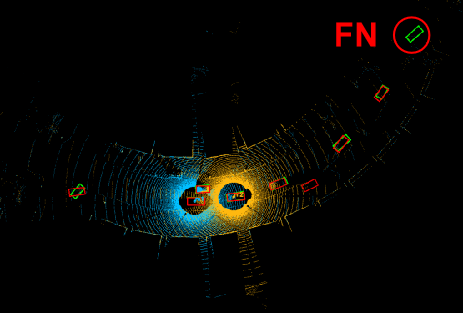}
    \subcaption{CoPEFT - Scene 1}
    \label{fig:vis_copeft_1}
  \end{minipage}
  \hfill
  \begin{minipage}[t]{0.48\linewidth}
    \centering
    \includegraphics[width=\linewidth]{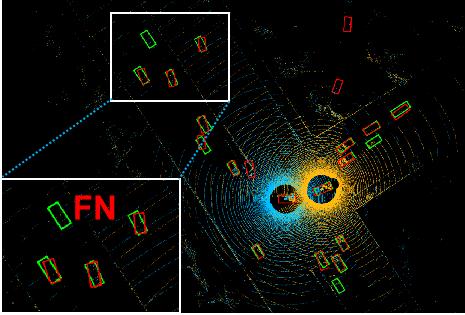}
    \subcaption{CoPEFT - Scene 2}
    \label{fig:vis_copeft_2}
  \end{minipage}

  \vspace{0.8em}

  % The thrid line
  \begin{minipage}[t]{0.48\linewidth}
    \centering
    \includegraphics[width=\linewidth]{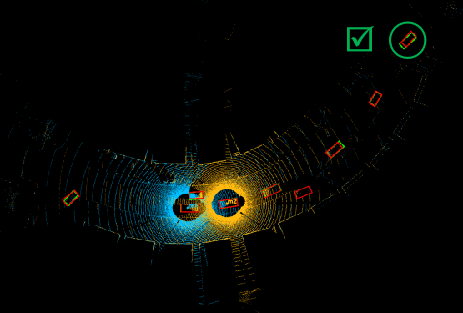}
    \subcaption{HeatV2X - Scene 1}
    \label{fig:vis_heatv2x_1}
  \end{minipage}
  \hfill
  \begin{minipage}[t]{0.48\linewidth}
    \centering
    \includegraphics[width=\linewidth]{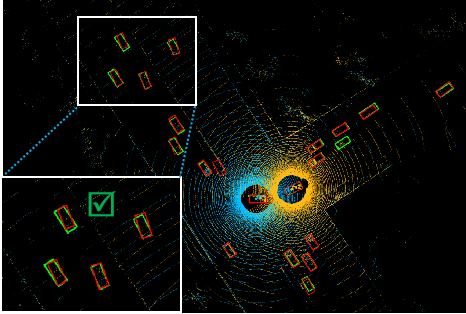}
    \subcaption{HeatV2X - Scene 2}
    \label{fig:vis_heatv2x_2}
  \end{minipage}

  \caption{\textbf{Visualization comparison across different scenes.}
  Predictions and ground truths are shown in \textcolor{red}{red} and \textcolor{green}{green}, respectively.
  HeatV2X achieves more accurate perception and successfully captures previously False Negative (FN) objects through efficient alignment and enhanced collaboration.}
  \label{fig:vis_compare}
\end{figure}

\cref{tab:mainresults} and~\cref{tab:dairv2x_results} compares detection performance and trainable parameters. Our collaborative framework achieves superior perception accuracy compared with the latest collaborative perception methods, while significantly reducing the number of trainable parameters relative to state-of-the-art end-to-end training approaches. We visualize the detection results of HEAL, CoPEFT, and our method. As shown in~\cref{fig:vis_compare}, our HeatV2X successfully captures objects missed by the other two methods and provides more accurate detections. This improvement is likely due to the integration of LHFT and GCFT methods. Specifically, LHFT enables rapid alignment within a unified representation space, while GCFT further captures cross-agent interactions to enhance collaborative perception.

Our approach achieves a 3\% improvement in AP@0.5 over state-of-the-art methods on the OPV2V-H dataset and further surpasses existing approaches across both AP@0.5 and AP@0.7 metrics on the DAIR-V2X dataset. Furthermore, through the fine-tuning of structured adapters, heterogeneous models can be aligned with less than $1 M$ parameters, leading to a consistent improvement in overall cooperative perception performance.

\subsubsection{Heterogeneous Collaborative Performance}
The reverse alignment significantly enhances the overall performance of collaborative perception, primarily due to the introduction of viewpoint and modality compensation from new agents. This raises an important question: \textbf{Could the collaboration framework itself become more efficient with the participation of heterogeneous agents?} To validate the ability of GCFT to strengthen the representation of heterogeneous collaboration, we conducted comparative experiments, which is shown in \cref{tab:hmca_finetuning}. First, we evaluated the performance of model $m_1$ before and after GCFT under single-vehicle perception, observing negligible differences. Then, under homogeneous-agent collaboration based on $m_1$, GCFT yielded a remarkable improvement. This confirms that GCFT does not enhance individual perception performance, but rather boosts the model by exploiting collaborative perception potential. This shows that collaborative compensation constitutes an additional mechanism for performance enhancement, alongside viewpoint and modality compensation.

\begin{table}[htbp]
\centering
\caption{Performance comparison before and after Global Collaborative Fine-Tuning (GCFT), where $A_x$ denotes different Agents and $m_x$ represents different modalities. Evaluation uses single-vehicle perception ground truth for single-Agent scenarios and joint perception ground truth for multi-Agent scenarios.}
\label{tab:hmca_finetuning}
\setlength{\tabcolsep}{3.3pt}
\begin{tabular*}{\columnwidth}{@{\extracolsep{\fill}}cccccccc@{}}
\toprule
\multicolumn{4}{c}{Model in Agent} & \multicolumn{2}{c}{Before GCFT} & \multicolumn{2}{c}{After GCFT} \\
\cmidrule(lr){1-4} \cmidrule(lr){5-6} \cmidrule(lr){7-8}
$A_1$ & $A_2$ & $A_3$ & $A_4$ & AP0.5 & AP0.7 & AP0.5 & AP0.7 \\
\midrule
$m_1$ & - & - & - & 0.786 & 0.662 & 0.788 & 0.665 \\
$m_1$ & $m_1$ & $m_1$ & $m_1$ & 0.924 & 0.809 & \textbf{0.958} & \textbf{0.879} \\
$m_1$ & $m_2$ & $m_3$ & $m_4$ & 0.878 & 0.733 & \textbf{0.914} & \textbf{0.806} \\
\bottomrule
\end{tabular*}
\end{table}

\subsubsection{Extended Experiments}
\textbf{Scalability Evaluation.} We compare trainable parameters and total GPU training time as the number of agents increases from 0 to 10. All trainable components, including pre-training, are included in the parameter statistics. The sharp performance rise of HEAL and HeatV2X from zero to one agent corresponds to their pre-training phase, after which the growth slows as further agents require no separate pre-training. Conversely, STAMP pre-trains every new agent, maintaining a consistently high growth trend (\cref{fig:hetero_and_scalable}). Experimental results demonstrate that even an untrained new agent can achieve efficient collaborative alignment with fewer trainable parameters and reduced training time using our method.

\begin{figure}[!htbp]
\centering
\includegraphics[width=3.3in]{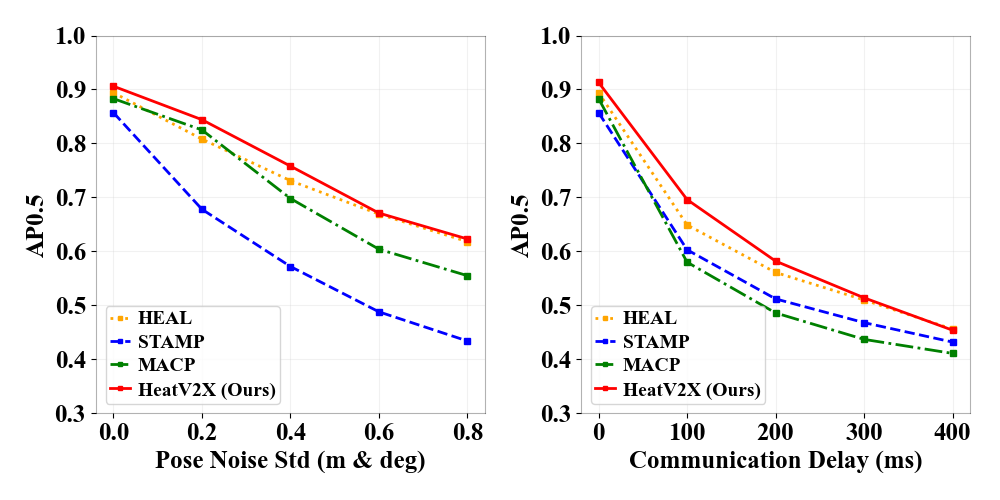}
\caption{The comparison of Robustness on the pose error and time delay setting.}
\label{fig:robust}
\end{figure}

\textbf{Robustness Evaluation}. To further compare perception performance under different noise conditions, we evaluate HeatV2X against the baseline under multiple noise levels. Specifically, we inject Gaussian noise with variances ranging from 0 to 0.8 into the OPV2V-H dataset and focus on the performance of four-agent collaborative perception. As shown in \cref{fig:robust}, our method maintains strong performance under high noise and large latency conditions, outperforming the baseline methods.

\subsection{Ablation Study}
Additionally, we conducted ablation studies to evaluate the performance and effectiveness of the LHFT and GCFT strategies. The ablation results on the OPV2V-H dataset are summarized in the \cref{tab:ablation}. As observed, incorporating LHFT consistently improves perception performance while reducing the number of trainable parameters. Furthermore, the addition of GCFT leads to further performance gains. These results confirm the effectiveness and complementary nature of each component and strategy in the HeatV2X.

\begin{table}[htbp]
\centering
\caption{\textbf{Ablation study results on OPV2V with 4 agents.} The Trainable Params refer to the total number of parameters involved in training, including 6.61M from the pre-training stage.}
\label{tab:ablation}
\setlength{\tabcolsep}{4pt} % 调整列间距 
\begin{tabular}{lccc}
\toprule
\textbf{Module} & \textbf{AP0.5} & \textbf{AP0.7} & \textbf{Trainable Params (M)} \\
\midrule
- & 0.852 & 0.713 & 54.59 \\
+LHFT & 0.872 & 0.716 & 2.34 (+6.61) \\
+GCFT & 0.876 & 0.709 & 0.13 (+6.61) \\
+LHFT+GCFT & \textbf{0.909} & \textbf{0.793} & 2.47 (+6.61) \\
\bottomrule
\end{tabular}
\end{table}

\begin{figure}[!t]
\centering
\includegraphics[width=3.2in]{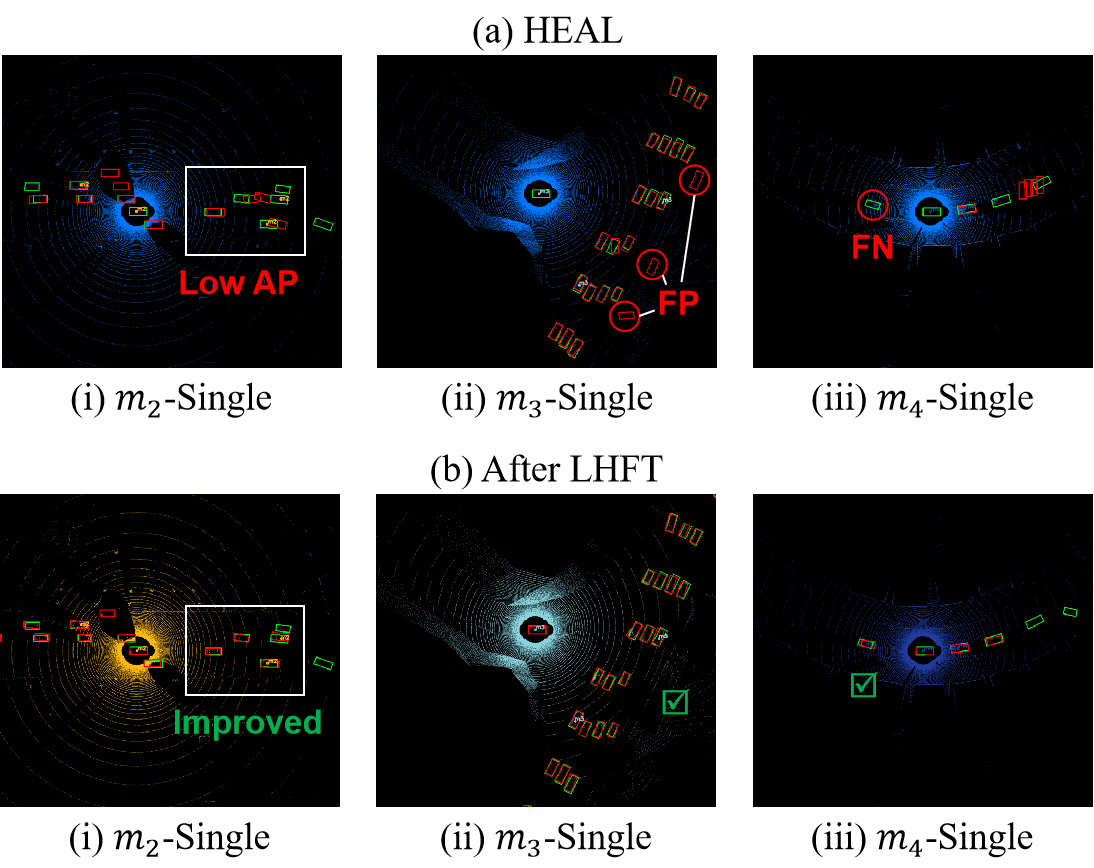}
\caption{\textbf{Visualization of feature-aligned outputs.} The results demonstrate that our LHFT strategy surpasses HEAL’s single-agent performance with fewer trainable parameters.}
\label{fig:vis_ablation_lhft}
\end{figure}

\begin{figure}[!htbp]
  \centering
  % The first line
  \begin{minipage}[t]{0.48\linewidth}
    \centering
    \includegraphics[width=\linewidth]{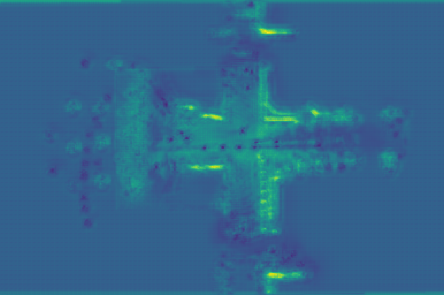}
    \subcaption{Heatmap (Before GCFT)}
    \label{fig:heatmap_before_gcft}
  \end{minipage}
  \hfill
  \begin{minipage}[t]{0.48\linewidth}
    \centering
    \includegraphics[width=\linewidth]{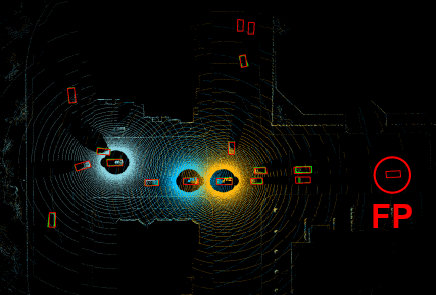}
    \subcaption{Detection (Before GCFT)}
    \label{fig:det_before_gcft}
  \end{minipage}

  \vspace{0.8em}

  % The second line
  \begin{minipage}[t]{0.48\linewidth}
    \centering
    \includegraphics[width=\linewidth]{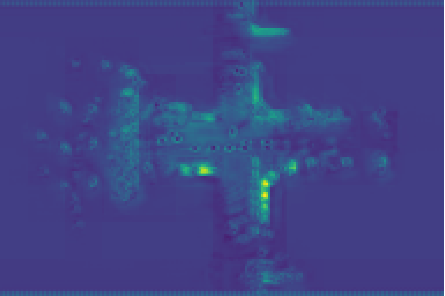}
    \subcaption{Heatmap (After GCFT)}
    \label{fig:heatmap_after_gcft}
  \end{minipage}
  \hfill
  \begin{minipage}[t]{0.48\linewidth}
    \centering
    \includegraphics[width=\linewidth]{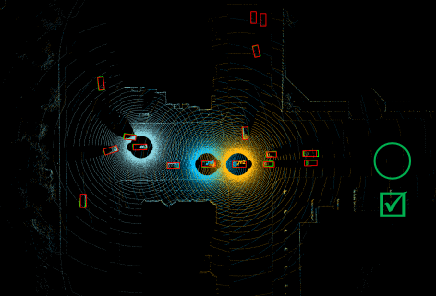}
    \subcaption{Detection (After GCFT)}
    \label{fig:det_after_gcft}
  \end{minipage}

  \vspace{0.8em}

  \caption{\textbf{Visualization comparison before and after GCFT.} After applying GCFT, the feature heatmaps become clearer and the previously False Positive (FP) objects are eliminated.}
  \label{fig:vis_ablation_gcft}
\end{figure}

% We further visualize the ablation results of LHFT and GCFT.
\textbf{Visualization.} Given that both HEAL’s backward alignment and our LHFT are designed to align untrained agents, we visualize their results in~\cref{fig:vis_ablation_lhft}. The results show that the single-agent performance after applying LHFT not only maintains low parameter overhead but also surpasses that of HEAL itself. In addition, as shown in~\cref{fig:vis_ablation_gcft}, the GCFT ablation experiment demonstrates that the MC Adapter effectively enhances inter-agent collaboration, successfully eliminating false detections observed before GCFT.

% \subsection{Visualization}

\section{Conclusion}
\label{sec:Conclusion}
In this paper, we propose Heterogeneous Adaptation (HeatV2X), a scalable collaborative framework based on fine-tuning. By performing Local Heterogeneous Fine-Tuning (LHFT) and Global Collaborative Fine-Tuning (GCFT), HeatV2X enriches heterogeneous information and exploits the potential of collaborative interaction at a low parameter cost. Experiments on the OPV2V-H and DAIR-V2X datasets show that our method achieves superior advantages in both performance and parameter efficiency over current state-of-the-art methods. In future work, we will further explore the integration of our collaborative framework with end-to-end autonomous driving and vision-language-action (VLA).

% It demonstrates that the collaborative framework can enhance overall perception performance by deepening inter-agent interaction without altering the individual agent’s perception ability.

{
    \small
    \bibliographystyle{ieeenat_fullname}
    \bibliography{main}

@String(CVPR= {IEEE Conf. Comput. Vis. Pattern Recog.})

@String(ICCV= {Int. Conf. Comput. Vis.})

@String(ECCV= {Eur. Conf. Comput. Vis.})

@String(AAAI = {AAAI})

@String(CVPR  = {CVPR})

@String(ICCV  = {ICCV})

@String(ECCV  = {ECCV})

@inproceedings{
lu2024an,
title={An Extensible Framework for Open Heterogeneous Collaborative Perception},
author={Lu, Yifan and Hu, Yue and Zhong, Yiqi and Wang, Dequan and Chen, Siheng and Wang, Yanfeng},
booktitle={The Twelfth International Conference on Learning Representations},
year={2024},
}

@inproceedings{ma2024macp,
  title={MACP: Efficient Model Adaptation for Cooperative Perception},
  author={Ma, Yunsheng and Lu, Juanwu and Cui, Can and Zhao, Sicheng and Cao, Xu and Ye, Wenqian and Wang, Ziran},
  booktitle={Proceedings of the IEEE/CVF Winter Conference on Applications of Computer Vision},
  pages={3373--3382},
  year={2024}
}

@INPROCEEDINGS{qu2023sicp,
  author={Qu, Deyuan and Chen, Qi and Bai, Tianyu and Lu, Hongsheng and Fan, Heng and Zhang, Hao and Fu, Song and Yang, Qing},
  booktitle={2024 IEEE/RSJ International Conference on Intelligent Robots and Systems (IROS)}, 
  title={SiCP: Simultaneous Individual and Cooperative Perception for 3D Object Detection in Connected and Automated Vehicles}, 
  year={2024},
  volume={},
  number={},
  pages={8905-8912},
  doi={10.1109/IROS58592.2024.10801398}
}

@InProceedings{Xiang_2023_ICCV,
    author    = {Xiang, Hao and Xu, Runsheng and Ma, Jiaqi},
    title     = {HM-ViT: Hetero-Modal Vehicle-to-Vehicle Cooperative Perception with Vision Transformer},
    booktitle = {Proceedings of the IEEE/CVF International Conference on Computer Vision (ICCV)},
    month     = {October},
    year      = {2023},
    pages     = {284-295}
}

@ARTICLE{10791908,
  author={Li, Zhiqi and Wang, Wenhai and Li, Hongyang and Xie, Enze and Sima, Chonghao and Lu, Tong and Yu, Qiao and Dai, Jifeng},
  journal={IEEE Transactions on Pattern Analysis and Machine Intelligence}, 
  title={BEVFormer: Learning Bird’s-Eye-View Representation From LiDAR-Camera via Spatiotemporal Transformers}, 
  year={2025},
  volume={47},
  number={3},
  pages={2020-2036},
  doi={10.1109/TPAMI.2024.3515454}
}

@inproceedings{wei2025copeft,
  title={CoPEFT: Fast Adaptation Framework for Multi-Agent Collaborative Perception with Parameter-Efficient Fine-Tuning},
  author={Wei, Quanmin and Dai, Penglin and Li, Wei and Liu, Bingyi and Wu, Xiao},
  booktitle={Proceedings of the AAAI Conference on Artificial Intelligence},
  volume={39},
  number={22},
  pages={23351--23359},
  year={2025}
}

@inproceedings{yin20255,
  title={5\%> 100\%: Breaking performance shackles of full fine-tuning on visual recognition tasks},
  author={Yin, Dongshuo and Hu, Leiyi and Li, Bin and Zhang, Youqun and Yang, Xue},
  booktitle={Proceedings of the Computer Vision and Pattern Recognition Conference},
  pages={20071--20081},
  year={2025}
}

@article{gao2025stamp,
  title={Stamp: Scalable task and model-agnostic collaborative perception},
  author={Gao, Xiangbo and Xu, Runsheng and Li, Jiachen and Wang, Ziran and Fan, Zhiwen and Tu, Zhengzhong},
  journal={arXiv preprint arXiv:2501.18616},
  year={2025}
}

@ARTICLE{10979246,
  author={Liu, Genjia and Hu, Yue and Xu, Chenxin and Mao, Weibo and Ge, Junhao and Huang, Zhengxiang and Lu, Yifan and Xu, Yinda and Xia, Junkai and Wang, Yafei and Chen, Siheng},
  journal={IEEE Transactions on Pattern Analysis and Machine Intelligence}, 
  title={Toward Collaborative Autonomous Driving: Simulation Platform and End-to-End System}, 
  year={2025},
  volume={47},
  number={8},
  pages={6566-6584},
  doi={10.1109/TPAMI.2025.3560327}
}

@InProceedings{Gao_2025_CVPR,
    author    = {Gao, Xiangbo and Wu, Yuheng and Wang, Rujia and Liu, Chenxi and Zhou, Yang and Tu, Zhengzhong},
    title     = {LangCoop: Collaborative Driving with Language},
    booktitle = {Proceedings of the IEEE/CVF Conference on Computer Vision and Pattern Recognition (CVPR) Workshops},
    month     = {June},
    year      = {2025},
    pages     = {4235-4246}
}

@inproceedings{xu2022v2xvit,
 author = {Runsheng Xu and Hao Xiang and Zhengzhong Tu and Xin Xia and Ming-Hsuan Yang and Jiaqi Ma},
 title = {V2X-ViT: Vehicle-to-Everything Cooperative Perception with Vision Transformer},
 booktitle={Proceedings of the European Conference on Computer Vision (ECCV)},
 year = {2022}
}

@inproceedings{dair-v2x,
  title={Dair-v2x: A large-scale dataset for vehicle-infrastructure cooperative 3d object detection},
  author={Yu, Haibao and Luo, Yizhen and Shu, Mao and Huo, Yiyi and Yang, Zebang and Shi, Yifeng and Guo, Zhenglong and Li, Hanyu and Hu, Xing and Yuan, Jirui and Nie, Zaiqing},
  booktitle={Proceedings of the IEEE/CVF Conference on Computer Vision and Pattern Recognition},
  pages={21361--21370},
  year={2022}
}

@inproceedings{Where2comm:22,
  author    = {Yue Hu and Shaoheng Fang and Zixing Lei and Yiqi Zhong and Siheng Chen},
  title     = {Where2comm: Communication-Efficient Collaborative Perception via Spatial Confidence Maps},
  booktitle = {Thirty-sixth Conference on Neural Information Processing Systems (Neurips)},
  month     = {November},  
  year      = {2022}
}

@InProceedings{V2VNet,
author="Wang, Tsun-Hsuan
and Manivasagam, Sivabalan
and Liang, Ming
and Yang, Bin
and Zeng, Wenyuan
and Urtasun, Raquel",
editor="Vedaldi, Andrea
and Bischof, Horst
and Brox, Thomas
and Frahm, Jan-Michael",
title="V2VNet: Vehicle-to-Vehicle Communication for Joint Perception and Prediction",
booktitle="Computer Vision -- ECCV 2020",
year="2020",
publisher="Springer International Publishing",
address="Cham",
pages="605--621",
isbn="978-3-030-58536-5"
}

@inproceedings{NEURIPS2021_f702defb,
 author = {Li, Yiming and Ren, Shunli and Wu, Pengxiang and Chen, Siheng and Feng, Chen and Zhang, Wenjun},
 booktitle = {Advances in Neural Information Processing Systems},
 editor = {M. Ranzato and A. Beygelzimer and Y. Dauphin and P.S. Liang and J. Wortman Vaughan},
 pages = {29541--29552},
 publisher = {Curran Associates, Inc.},
 title = {Learning Distilled Collaboration Graph for Multi-Agent Perception},
 url = {https://proceedings.neurips.cc/paper_files/paper/2021/file/f702defbc67edb455949f46babab0c18-Paper.pdf},
 volume = {34},
 year = {2021}
}

@InProceedings{Zhang_2024_CVPR,
  author    = {Zhang, Jingyu and Yang, Kun and Wang, Yilei and Wang, Hanqi and Sun, Peng and Song, Liang},
  title     = {ERMVP: Communication-Efficient and Collaboration-Robust Multi-Vehicle Perception in Challenging Environments},
  booktitle = {Proceedings of the IEEE/CVF Conference on Computer Vision and Pattern Recognition (CVPR)},
  month     = {June},
  year      = {2024},
  pages     = {12575-12584}
}

@inproceedings{xu2022cobevt,
 author = {Runsheng Xu and Zhengzhong Tu and Hao Xiang and  Wei Shao and Bolei Zhou and Jiaqi Ma},
 title = {CoBEVT: Cooperative Bird's Eye View Semantic Segmentation with Sparse Transformers},
 booktitle={Conference on Robot Learning (CoRL)},
 year = {2022}
}

@ARTICLE{10265751,
  author={Yin, Hongbo and Tian, Daxin and Lin, Chunmian and Duan, Xuting and Zhou, Jianshan and Zhao, Dezong and Cao, Dongpu},
  journal={IEEE Transactions on Intelligent Transportation Systems}, 
  title={V2VFormer++: Multi-Modal Vehicle-to-Vehicle Cooperative Perception via Global-Local Transformer}, 
  year={2024},
  volume={25},
  number={2},
  pages={2153-2166},
  doi={10.1109/TITS.2023.3314919}
}

@InProceedings{Yuan_2025_CVPR,
    author    = {Yuan, Yunshuang and Xia, Yan and Cremers, Daniel and Sester, Monika},
    title     = {SparseAlign: a Fully Sparse Framework for Cooperative Object Detection},
    booktitle = {Proceedings of the IEEE/CVF Conference on Computer Vision and Pattern Recognition (CVPR)},
    month     = {June},
    year      = {2025},
    pages     = {22296-22305}
}

@InProceedings{pnpda,
author="Luo, Tianyou
and Yuan, Quan
and Luo, Guiyang
and Xia, Yuchen
and Yang, Yujia
and Li, Jinglin",
title="Plug and Play: A Representation Enhanced Domain Adapter for Collaborative Perception",
booktitle="Computer Vision -- ECCV 2024",
year="2025",
}

@inproceedings{xu2023mpda,
 author = {Runsheng Xu and Jinlong Li and Xiaoyu Dong and Hongkai Yu and Jiaqi Ma},
 title = {Bridging the Domain Gap for Multi-Agent Perception},
 booktitle={2023 IEEE International Conference on Robotics and Automation (ICRA)},
 year = {2023}
}

@inproceedings{hu2022lora,
title={Lo{RA}: Low-Rank Adaptation of Large Language Models},
author={Edward J Hu and Yelong Shen and Phillip Wallis and Zeyuan Allen-Zhu and Yuanzhi Li and Shean Wang and Lu Wang and Weizhu Chen},
booktitle={International Conference on Learning Representations},
year={2022},
url={https://openreview.net/forum?id=nZeVKeeFYf9}
}

@inproceedings{bitfit ,
title = {BitFit: Simple Parameter-efficient Fine-tuning for Transformer-based Masked Language-models},
journal = {Proceedings of the Annual Meeting of the Association for Computational Linguistics},
author = {Ben-Zaken, Elad and Ravfogel, Shauli and Goldberg, Yoav},
year = {2022},
URL = {http://dx.doi.org/10.18653/v1/2022.acl-short.1},
}

@InProceedings{adapter,
  title = 	 {Parameter-Efficient Transfer Learning for {NLP}},
  author =       {Houlsby, Neil and Giurgiu, Andrei and Jastrzebski, Stanislaw and Morrone, Bruna and De Laroussilhe, Quentin and Gesmundo, Andrea and Attariyan, Mona and Gelly, Sylvain},
  booktitle = 	 {Proceedings of the 36th International Conference on Machine Learning},
  pages = 	 {2790--2799},
  year = 	 {2019},
  volume = 	 {97},
  series = 	 {Proceedings of Machine Learning Research},
  month = 	 {09--15 Jun},
  publisher =    {PMLR},
  url = 	 {https://proceedings.mlr.press/v97/houlsby19a.html},
}

@article{Xu2021OPV2VAO,
  title={OPV2V: An Open Benchmark Dataset and Fusion Pipeline for Perception with Vehicle-to-Vehicle Communication},
  author={Runsheng Xu and Hao Xiang and Xin Xia and Xu Han and Jinlong Liu and Jiaqi Ma},
  journal={2022 International Conference on Robotics and Automation (ICRA)},
  year={2021},
  pages={2583-2589},
  url={https://api.semanticscholar.org/CorpusID:237532732}
}

@misc{han2024peft,
      title={Parameter-Efficient Fine-Tuning for Large Models: A Comprehensive Survey}, 
      author={Zeyu Han and Chao Gao and Jinyang Liu and Jeff Zhang and Sai Qian Zhang},
      year={2024},
      eprint={2403.14608},
      archivePrefix={arXiv},
      primaryClass={cs.LG},
      url={https://arxiv.org/abs/2403.14608}, 
}

@article{zhou2025pragmatic,
  title={Pragmatic Heterogeneous Collaborative Perception via Generative Communication Mechanism},
  author={Zhou, Junfei and Dai, Penglin and Wei, Quanmin and Liu, Bingyi and Wu, Xiao and Wang, Jianping},
  journal={arXiv preprint arXiv:2510.19618},
  year={2025}
}

@inproceedings{lang2019pointpillars,
  title={Pointpillars: Fast encoders for object detection from point clouds},
  author={Lang, Alex H and Vora, Sourabh and Caesar, Holger and Zhou, Lubing and Yang, Jiong and Beijbom, Oscar},
  booktitle={Proceedings of the IEEE/CVF conference on computer vision and pattern recognition},
  pages={12697--12705},
  year={2019}
}

@Article{yan2018second,
AUTHOR = {Yan, Yan and Mao, Yuxing and Li, Bo},
TITLE = {SECOND: Sparsely Embedded Convolutional Detection},
JOURNAL = {Sensors},
VOLUME = {18},
YEAR = {2018},
NUMBER = {10},
ARTICLE-NUMBER = {3337},
PubMedID = {30301196},
ISSN = {1424-8220},
DOI = {10.3390/s18103337}
}

@inproceedings{philion2020lift,
    title={Lift, Splat, Shoot: Encoding Images From Arbitrary Camera Rigs by Implicitly Unprojecting to 3D},
    author={Jonah Philion and Sanja Fidler},
    booktitle={Proceedings of the European Conference on Computer Vision},
    year={2020},
}

@InProceedings{pmlr-v97-tan19a,
  title = 	 {{E}fficient{N}et: Rethinking Model Scaling for Convolutional Neural Networks},
  author =       {Tan, Mingxing and Le, Quoc},
  booktitle = 	 {Proceedings of the 36th International Conference on Machine Learning},
  pages = 	 {6105--6114},
  year = 	 {2019},
  editor = 	 {Chaudhuri, Kamalika and Salakhutdinov, Ruslan},
  volume = 	 {97},
  series = 	 {Proceedings of Machine Learning Research},
  month = 	 {09--15 Jun},
  publisher =    {PMLR},
}

@INPROCEEDINGS{resnet,
  author={He, Kaiming and Zhang, Xiangyu and Ren, Shaoqing and Sun, Jian},
  booktitle={2016 IEEE Conference on Computer Vision and Pattern Recognition (CVPR)}, 
  title={Deep Residual Learning for Image Recognition}, 
  year={2016},
  volume={},
  number={},
  pages={770-778},
  doi={10.1109/CVPR.2016.90}
}

@article{li2024delving,
  title={Delving into the secrets of BEV 3D object detection in autonomous driving: a comprehensive survey},
  author={Li, Haoyu and Zhao, Yueran and Zhong, Jiaru and Wang, Bo and Sun, Chao and Sun, Fengchun},
  journal={Authorea Preprints},
  year={2024},
  publisher={Authorea}
}

@article{zhang2025height3d,
  title={Height3d: A roadside visual framework based on height prediction in real 3-d space},
  author={Zhang, Zhang and Sun, Chao and Wang, Bo and Guo, Bin and Wen, Da and Zhu, Tianyi and Ning, Qili},
  journal={IEEE Transactions on Intelligent Transportation Systems},
  year={2025},
  publisher={IEEE}
}

@inproceedings{YueCodeFilling:CVPR2024,
 author = {Yue Hu and Juntong Peng and Sifei Liu and Junhao Ge and Si Liu and Siheng Chen},
 title = {Communication-Efficient Collaborative Perception via Information Filling with Codebook},
 booktitle = {2024 IEEE / CVF Computer Vision and Pattern Recognition Conference (CVPR)},
 year = {2024}
}

@inproceedings{yu2024_univ2x,
 title={End-to-End Autonomous Driving through V2X Cooperation}, 
 author={Haibao Yu and Wenxian Yang and Jiaru Zhong and Zhenwei Yang and Siqi Fan and Ping Luo and Zaiqing Nie},
 booktitle={The 39th Annual AAAI Conference on Artificial Intelligence},
 year={2025}
}

@article{zhang2025heightformer,
  title={Heightformer: Learning height prediction in voxel features for roadside vision centric 3d object detection via transformer},
  author={Zhang, Zhang and Sun, Chao and Yue, Chao and Wen, Da and Chen, Yujie and Wang, Tianze and Leng, Jianghao},
  journal={arXiv preprint arXiv:2503.10777},
  year={2025}
}

@article{zhang2025pillarmamba,
  title={Pillarmamba: Learning local-global context for roadside point cloud via hybrid state space model},
  author={Zhang, Zhang and Sun, Chao and Yue, Chao and Wen, Da and Wang, Tianze and Leng, Jianghao},
  journal={arXiv preprint arXiv:2505.05397},
  year={2025}
}

@inproceedings{Dosovitskiy17,
  title = {{CARLA}: {An} Open Urban Driving Simulator},
  author = {Alexey Dosovitskiy and German Ros and Felipe Codevilla and Antonio Lopez and Vladlen Koltun},
  booktitle = {Proceedings of the 1st Annual Conference on Robot Learning},
  pages = {1--16},
  year = {2017}
}
}
% \input{sec/6_appendix}

\end{document}